\documentclass[sigconf,nonacm,arxiv]{acmart}

\usepackage{multirow}
\usepackage{algorithm}
\usepackage{algorithmic}
\usepackage{graphicx}
\usepackage{verbatim}

\usepackage{etoolbox}
\makeatletter
\g@addto@macro\verbatim\footnotesize
\makeatother

\usepackage{geometry}
\AtBeginDocument{%
  \providecommand\BibTeX{{%
    \normalfont B\kern-0.5em{\scshape i\kern-0.25em b}\kern-0.8em\TeX}}}

\setcopyright{acmcopyright}
\copyrightyear{2023}
\acmYear{2023}
\acmDOI{XXXXXXX.XXXXXXX}

\acmConference[Conference acronym 'XX]{Make sure to enter the correct
  conference title from your rights confirmation emai}{June 03--05,
  2018}{Woodstock, NY}
\acmPrice{15.00}
\acmISBN{978-1-4503-XXXX-X/18/06}

\begin{document}


\title{Auto-GPT for Online Decision Making: Benchmarks and Additional Opinions}



\author{Hui Yang}
\affiliation{%
  \institution{Amazon}
  \country{United States}
}
\email{younghumanfly@gmail.com}

\author{Sifu Yue}
\affiliation{%
  \institution{Amazon}
  \country{United States}
}
\email{eddieyue1018@gmail.com}

\author{Yunzhong He}
\affiliation{%
  \institution{Kaito.ai}
  \country{United States}
}
\email{webroot@kaito.ai}

\renewcommand{\shortauthors}{}

\begin{abstract}

Auto-GPT is an autonomous agent that leverages recent advancements in adapting Large Language Models (LLMs) for decision-making tasks. While there has been a growing interest in Auto-GPT stypled agents, questions remain regarding the effectiveness and flexibility of Auto-GPT in solving real-world decision-making tasks. Its limited capability for real-world engagement and the absence of benchmarks contribute to these uncertainties. In this paper, we present a comprehensive benchmark study of Auto-GPT styled agents in decision-making tasks that simulate real-world scenarios. Our aim is to gain deeper insights into this problem and understand the adaptability of GPT-based agents. We compare the performance of popular LLMs such as GPT-4, GPT-3.5, Claude, and Vicuna in Auto-GPT styled decision-making tasks. Furthermore, we introduce the Additional Opinions algorithm, an easy and effective method that incorporates supervised/imitation-based learners into the Auto-GPT scheme. This approach enables lightweight supervised learning without requiring fine-tuning of the foundational LLMs. We demonstrate through careful baseline comparisons and ablation studies that the Additional Opinions algorithm significantly enhances performance in online decision-making benchmarks, including WebShop and ALFWorld. 
\end{abstract}

\maketitle

\section{Introduction}\label{sec:intro}
Adapting Large Language Models (LLM) to autonomous agents have recently achieved great success in various decision-making\cite{palm}, virtual character simulation\cite{generativeagent} and tool manipulation tasks\cite{taskmatrix}. While there are evidence suggesting that scaling up LLM can lead to certain level of general intelligence\cite{msgpt}, there are still limitations in leveraging LLM as autonomous agents directly. The lack of long-term memory, limited token length, and the lack of deterministic control over its behaviors, etc. are among them. Various techniques in prompting\cite{cot}, planning and memory retrieval\cite{cot, taskmatrix, generativeagent} to overcome the limitations are proposed recently and worked well. Among all of the agents, there is Auto-GPT\footnote{https://github.com/Significant-Gravitas/Auto-GPT}, a GPT-based autonomous agent that connects to the internet and attempts to accomplish any task. Despite its sudden surge in attention, the extent of its effectiveness in accomplishing tasks remains uncertain due to its limited capacity for action.

We characterize Auto-GPT styled agent with the following properties (1) it receives only high-level goals and instructions at the beginning of complex multi-step tasks, without requiring step-by-step guidance from humans (2) it engages in self-monologue by generating 'Thoughts,' 'Reasoning,' 'Plan,' and 'Criticism' for each individual step of action (effectively, CoT and Reflexion \cite{cot, reflexion}) (3) it possesses the capability to integrate various tools through simple tool instructions and a few examples (4) it incorporates long-term self-memory and memory retrieval mechanisms (5) tasks-specific adaption should only require minimal efforts like providing goal definitions and tool descriptions.

To gain deeper insights into the performance and limitations of Auto-GPT styled agents, we conduct experiments by adapting Auto-GPT for online decision-making tasks that involve responding to unknown external environments. We evaluate various popular LLMs across multiple online learning tasks and present our findings and insights. Additionally, we propose a novel approach to demonstrate how external models can be leveraged as providers for additional opinions. Despite the recent advancements in LLM research, such as self-consistency and group voting techniques \cite{selfconsistency, complexitybased}, as well as incorporating external expert models and APIs to enhance LLMs \cite{hugginggpt, taskmatrix}, it has not been reported that LLMs like GPT-4 can benefit from additional opinions in the same way humans do \cite{additional_opinion}. This intriguing discovery opens up possibilities for a new paradigm where smaller expert models can collaborate with LLMs.

In this work we offer the following contributions (1) to the best of our knowledge, we are the first to show Auto-GPT can be easily adapted to online decision-making tasks that closely resemble real-world scenarios. (2) We provide comprehensive benchmark comparisons between popular LLMs, including GPT-4 \cite{gpt4}, GPT-3.5, Claude \cite{claude}, and Vicuna \cite{vicuna}. We present our findings regarding the adaptation of these models for autonomous agents. (3) We demonstrate that incorporating a second opinion from a supervised learner can significantly enhance task performance, providing a low-cost way of introducing supervision signals into Auto-GPT styled autonomous agent without model fine-tuning.

\section{Methodology}\label{sec:method}
\subsection{Tasks and baseline models}
\subsubsection{WebShop}
Webshop \cite{webshop} is a simulated environment that replicates a web shopping experience by scraping 1,181,436 products from Amazon.com and hosting them on an isolated server. The environment offers agents a realistic action space, including options such as performing product searches, clicking on items, navigating back to previous pages, and making purchases. Equipped with an integrated search engine, the environment provides the shopping agent with real-time observations that mimic those from a web browser. The evaluation process involves determining whether the agent successfully purchases the intended product based on its description, where a success requires all matches on the product itself, attributes, options and price together.
We use the IL (Imitation Learning) method with a fine-tuned action policy component as the baseline model, and compare it with popular generative LLMs with Auto-GPT styled adaption towards this web shopping task.

\subsubsection{ALFWorld}
ALFWorld \cite{alfworld} is a ground-breaking research environment that harmonizes the sophisticated, task-oriented language understanding of the ALFRED \cite{alfred} dataset with the immersive interactive fiction of TextWorld \cite{textworld}. The ALFRED (Action Learning From Realistic Environments and Directives) benchmark offers a robust testing ground for models to learn to parse and carry out intricate tasks from language directives within a detailed, interactive 3D environment. Meanwhile, TextWorld serves as a dynamic learning playground for training and evaluating reinforcement learning agents on text-based games. By interweaving these two platforms, ALFWorld brings together the linguistic comprehension and decision-making challenges of text-based games with the physical interactions in a 3D environment, embodying a critical step towards melding natural language instructions with real-world physical interactions. The environment contains over 25,000 unique, procedurally-generated tasks across photorealistic settings in various areas such as kitchens, living rooms, and bedrooms. These tasks require complex problem-solving skills and a thorough understanding of both language and environment, creating an elevated benchmark for AI performance. As much as ALFWorld presents a challenging yet fertile testbed for reinforcement learning, natural language understanding, and interactive decision-making research, we also start the evaluation process with the DAgger \cite{ross2011reduction} IL (Imitation Learning) agent against the unseen dataset as the baseline. Then we benchmark it against prevailing generative language learning models that utilize an Auto-GPT style approach, with these models only being adjusted for the ALFWorld task with tool demonstrations.

\subsection{Prompt design}
We adapt Auto-GPT for both tasks without extensive tuning, simply by providing the task requirements or questions directly as Auto-GPT's goal. For instance, we input sentences such as \textit{"I want to purchase a folding storage box that is easy to install, made of faux leather, and has dimensions of 60x40x40cm"}. To facilitate Auto-GPT's understanding of available actions, we represent each action as a tool. It is worth noting that we observed poor performance when using tool instructions without examples in a sermon-style manner. However, with just a few examples, the performance improved significantly. Therefore, we include one to three few-shot examples for tool demonstrations, to harness the in-context learning abilities of LLMs.



\subsection{Considering additional opinions}
We further engineer changes to the Auto-GPT workflow to take additional opinions from external expert models into consideration. Specifically, we sample top k opinions from an expert model in Auto-GPT's decision phase, and present those opinions into the context section of the prompt towards more informed decisions. Details of the modified Auto-GPT workflow is outlined in Algorithm 1. In this work we simply use readily available IL models for both tasks as the external expert. The prompt to suggest additional opinions to LLM follows the template as '\textit{Here's one(a few) suggestion(s) for the command: <action with parameters> Please use this suggestion as a reference and make your own judgement.} '


\begin{figure}[hbt!] 
  \centering
  \includegraphics[width=\linewidth]{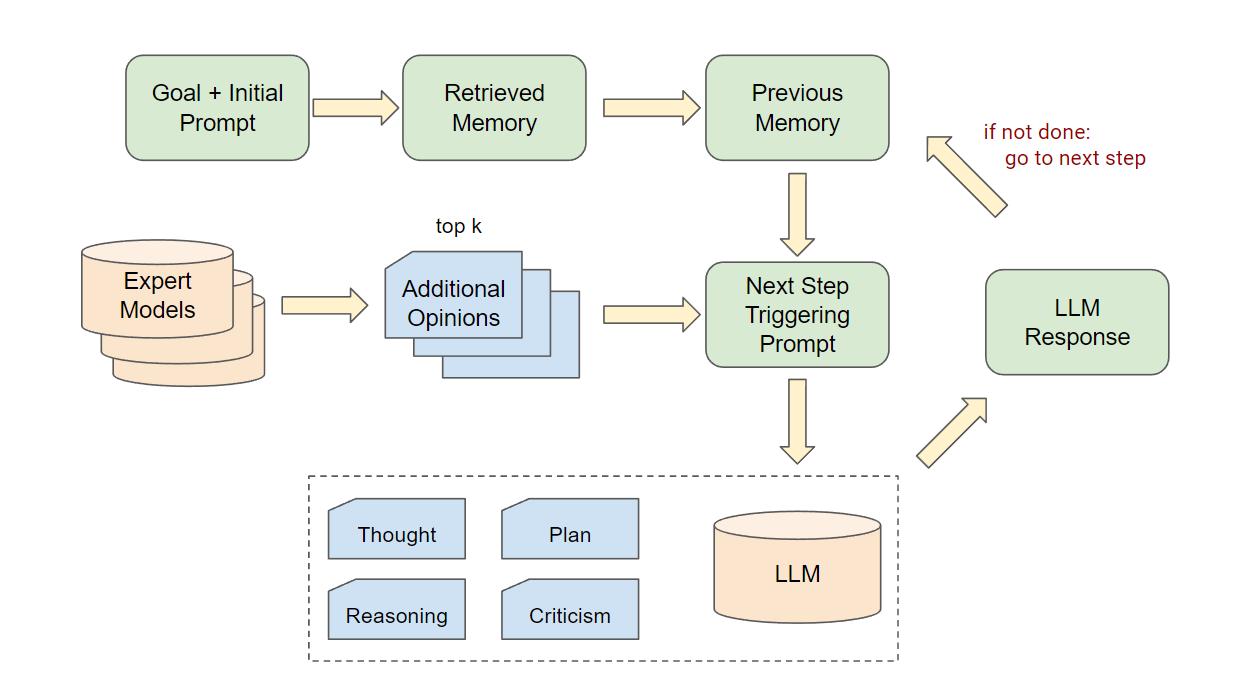}
  \caption{One step of Auto-GPT with Additional Opinions. Here Additional Opinions are generated by other expert models - IL models here, but extensible to any other models like Rule or other LLMs.}
  \label{fig:Auto-GPT_Additional_Opinion}
\end{figure}

\begin{algorithm}
\caption{Additional Opinion Auto-GPT Algorithm.}
\begin{algorithmic}[1] 
\REQUIRE
$o_i$: additional opinion sampled from expert model. 

$P_{o}$($o_i$): a prompt template wrapping top k $o_i$ as suggestions to LLM.
$P_{h}$: the regular prompt as a human to trigger LLM response.

Add(x): Add x into Auto-GPT context.

\STATE Initialize Auto-GPT
\FOR{each Auto-GPT step}
  \STATE Add(Initial Goal and Instruction Prompt)
  \IF{sampled $o_i$ from expert models exists}
    \STATE Add($P_{o}$($o_i$)) for i < k
  \ELSE
    \STATE Add($P_{h}$) 
  \ENDIF
  \STATE Auto-GPT runs with the prompt added.
\ENDFOR
\RETURN result
\end{algorithmic}
\end{algorithm}

\section{Experiments}\label{sec:offline_experiments}
\subsection{Experimental setup}

\begin{table*}[htbp]
  \centering
  \caption{Updated Webshop Model Performance Metrics}
  \label{tab:model_performance}
  \begin{tabular}{ccccc}
    \toprule
    \textbf{Model} & \textbf{Success Rate} & \textbf{Reward} & \textbf{Precision} & \textbf{Purchase Rate} \\
    \midrule
    \textbf{Base Models} & & & & \\
    Rule & 0.060 & 44.589 & 0.060 & 1.000 \\
    IL w/o. Image & 0.213 & 56.056 & 0.213 & 1.000 \\
    IL & 0.227 & 57.689 & 0.227 & 1.000 \\
    \midrule
    \textbf{Auto-GPT(Claude) Variants} & & & & \\
    Auto-GPT(Claude) & 0.140 & 47.617 & 0.146 & 0.960 \\
    Auto-GPT(Claude) + IL & 0.240 & 48.600 & 0.270 & 0.890 \\
    Auto-GPT(Claude) + IL(top5) & 0.220 & 52.010 & 0.229 & 0.960 \\
    \midrule
    \textbf{Auto-GPT(GPT3.5) Variants} & & & & \\
    Auto-GPT(GPT3.5) & 0.120 & 43.833 & 0.140 & 0.860 \\
    Auto-GPT(GPT3.5) + IL & 0.200 & 47.717 & 0.241 & 0.830 \\
    Auto-GPT(GPT3.5) + IL(top5) & 0.230 & 52.827 & 0.279 & 0.820 \\
    AutoGPT(GPT3.5) + Random & 0.060 & 22.333 & 0.136 & 0.440 \\
    \midrule
    \textbf{Auto-GPT(GPT4) Variants} & & & & \\
    Auto-GPT(GPT4) & 0.240 & 46.133 & 0.353 & 0.680 \\
    Auto-GPT(GPT4) + IL & 0.300 & 56.233 & 0.361 & 0.830 \\
    Auto-GPT(GPT4) + IL(top5) & \textbf{0.320} & \textbf{61.550} & \textbf{0.372} & 0.860 \\
    \midrule
    \textbf{Auto-GPT(Vicuna)} & & & & \\
    Auto-GPT(Vicuna) & 0.000 & 0.000 & 0.000 & 0.000 \\
    \bottomrule
  \end{tabular}
\end{table*}
\subsubsection{Webshop}
We utilized the original WebShop server setup from the GitHub Repository provided in the original paper\cite{webshop}. For testing, we adhered to a fixed order of iteration, selecting the first 50 instructions. This limited test set is a trade-off due to cost and computational efficiency concerns, especially running GPT4. Imitation Learning (IL) models, both with and without image embeddings, were used to ensure a fair comparison with large language models, given that the latter lack image access. The 'additional opinions' provided to Auto-GPT consistently used the superior IL model with image access. The temperature was set to 0.01 across all models to reduce randomness. Evaluation of IL models and Auto-GPT + IL variants followed a rigorous protocol to mitigate sampling randomness and assess small observed variations, respectively. In the case of Auto-GPT alone, we performed a single run due to the minimal variations observed as a result of the low temperature setting. Nonetheless, small variations were noticed in the Auto-GPT + IL variants, prompting us to conduct two runs and use their average for analysis. 

In the original research paper, a supervised Imitation Learning (IL) model was trained to aid the agent in making optimal decisions at each stage, with the ultimate goal of executing the correct purchase based on a given human instruction. The system was structured around four principal tasks: (1) generating a quality search query based on the original instruction, (2) selecting an item to browse from a variety of items and their titles, (3) deciding whether to check 'Description', 'Features' or 'Reviews' in the product detail page while also making the correct product option choices such as size, color etc., and (4) finalizing the purchase. 

Two IL models were utilized: a BART model for generating the most effective search query (task type 1) and a BERT model to make the correct choices for task type 2 to 4. Rule based model always directs to purcahse the first item after search.

\subsubsection{ALFWorld}
Adopting a similar approach to the Webshop experiment, we leverage the existing ALFWorld task simulation environment, including the unseen set of 134 games that was utilized for benchmarking in the original research paper. From the original study, we specifically incorporate the task agent (BUTLER::BRAIN) from the Imitation Learning (IL) model, excluding the visual input agent (BUTLER::VISION). The training of the text agent is performed using the DAgger (Dataset Aggregation) approach within an imitation learning context using expert demonstrations. To manage task execution failures, Beam Search is deployed to generate alternate action sentences, typically opting for the best sequence of words greedily for efficiency. The application of Beam Search primarily aims at enhancing the action sentence generation during failures rather than optimizing over embodied interactions. Echoing the Webshop experiment, to ensure a fair comparison with large language models (LLMs), we furnish the 'additional opinions' from the IL model to Auto-GPT. To control randomness, we maintain a temperature setting of 0.01 for all LLMs to minimize noise and strictly adhere to the original evaluation protocol to further mitigate randomness.

\begin{table*}[htbp]
  \centering
  \caption{Updated ALFWorld Model Performance Metrics}
  \label{tab:model_performance}
  \begin{tabular}{ccccc}
    \toprule
    \textbf{Model} & \textbf{Success Rate} & \textbf{Reward} & \textbf{Precision} & \textbf{Completion Rate} \\
    \midrule
    \textbf{Base Models, average of three} & & & & \\
    IL w/o. Beam Search & 0.179 & 24 & 1.000 & 0.179 \\
    IL & 0.306 & 41 & 1.000 & 0.441 \\
    \midrule
    \textbf{Auto-GPT(Claude) Variants} & & & & \\
    Auto-GPT(Claude) & 0.082 & 11 & 0.104 & 0.791 \\
    Auto-GPT(Claude) + IL & 0.090 & 12 & 0.130 & 0.687 \\
    \midrule
    \textbf{Auto-GPT(GPT3.5) Variants} & & & & \\
    Auto-GPT(GPT3.5) & 0.075 & 10 & 0.078 & 0.866 \\
    Auto-GPT(GPT3.5) + IL & 0.030 & 4 & 0.048 & 0.470 \\
    \midrule
    \textbf{Auto-GPT(GPT4) Variants} & & & & \\
    Auto-GPT(GPT4) & 0.485 & 65 & 0.628 & 0.582 \\
    Auto-GPT(GPT4) + IL & \textbf{0.515} & \textbf{69} & \textbf{0.789} & 0.530 \\
    \midrule
    \textbf{Auto-GPT(Vicuna)} & & & & \\
    Auto-GPT(Vicuna) & 0.000 & 0.000 & 0.000 & 0.000 \\
    \bottomrule
  \end{tabular}
\end{table*}

\subsection{Baseline comparison}
\subsubsection{Webshop}

Table 1 illustrates the results of running the first 50 test cases across the original IL models and Auto-GPT agent using different large language models (LLMs). The original IL model, lacking image input, only achieved a modest success rate, illustrating the complexity of the task. IL models incorporating image input as embeddings performed more favorably. Auto-GPT agents utilizing GPT3.5 or Claude alone performed worse than the original IL models, with or without images. However, GPT4 by itself exhibited superior performance compared to both IL models. 

A noteworthy point is that IL models display better rewards than Auto-GPT baselines without IL due to a higher purchase rate, and an allowance for more steps (100 vs. 20). However, the reward metric may not necessarily serve as the best end measurement, particularly considering real-world shopping scenarios where an agent refraining from making a purchase could be preferable to it making a purchase that doesn't entirely meet the requirements. In cases where a purchase is made, if we calculate precision solely based on these instances, Auto-GPT (GPT4), with or without IL, demonstrates significantly higher precision compared to any other variants. Further details on GPT4 considering the additional opinion can be found in Appendix 1.

\subsubsection{ALFWorld}
Table 2 presents the results of our ALFWorld experiment involving the Imitation Learning (IL) model and Large Language Models (LLMs) in an AutoGPT configuration, evaluated across an unseen set of 134 data points from ALFWorld's IL Model. Notably, the IL model with Beam Search significantly outperformed the version without Beam Search, as indicated by a considerable decrease in the success rate from 0.306 to 0.179 when Beam Search was omitted.

While Claude and GPT3.5 operating in the AutoGPT setting fell short of surpassing the IL model, GPT4 markedly exceeded the IL model's performance, irrespective of the use of Beam Search, despite the disatvantage of steps allowed (35 vs. 50). We hypothesize that the comparatively lower performance of Claude and GPT3.5 can be attributed to their lack of full episode demonstrations, and we posit that the introduction of more examples might enhance their performance.

This is particularly pertinent in tasks where the description carries implications, such as "heat the mug and put the mug on the countertop," which presupposes the agent's understanding of how to heat the mug. Both Claude and GPT3.5 struggled with deriving such implicit knowledge. However, we anticipate that supplementing the context with more examples would likely enhance their performance in such complex tasks.

\subsection{LLM comparison}
\subsubsection{Webshop}
Across all LLMs, Claude and GPT3.5 alone (0.140 vs. 0.120 success rate) compare similarly as each other in the AutoGPT setting. GPT4 alone performs the best (0.24 success rate) and Vicuna is tested not able to generate formatted responses thus reported 0 here. Another perspective to consider here is that the speed of calling these LLM APIs - Claude is faster than GPT3.5 and much faster than GPT4. We recommend here that Claude could be a great solution considering the performance latency tradeoff for real world problems.

\subsubsection{ALFWorld}
In line with our observations from the Webshop experiment, the success rates for Claude and GPT3.5, when applied in the AutoGPT setting, were relatively low, success rate at 0.075 and 0.082, respectively. Among the models, GPT4 demonstrated superior performance, achieving the highest success rate of 0.485 and a precision as high as 0.628, surpassing all other models including the Imitation Learning model. Despite Claude's advantage in speed over GPT3.5 and GPT4, its performance was markedly inferior to GPT4. Taking into account these observations, we recommend the use of GPT4 given its performance supremacy over the other models under consideration.

\subsection{Additional opinions}
\subsubsection{Webshop}
A novel paradigm emerged from our study, amalgamating Large Language Models (LLMs) with Expert models. Rather than solely relying on Expert models and their generated results, we propose an integrated approach. Firstly, the top k Additional Opinions are sampled from Expert models. Subsequently, these opinions are presented to the LLMs, prompting them to consider these views and make the final decision. This methodology particularly resonates with GPT4. Even though the underlying mechanism is still elusive and it is not yet clear if it mimics human decision-making processes\cite{additional_opinion}, the effectiveness of this approach is tangible in the experimental outcomes. Our working hypothesis postulates that GPT4 exhibits inherent biases when making autonomous decisions. However, by introducing opinions from various weak learners, GPT4 can enhance its performance. Considering these diverse viewpoints may allow GPT4 to mitigate its own biases and overcome inherent limitations. 

Interestingly, the inclusion of a single IL choice as an additional opinion in the context resulted in improved performance for all LLMs. This performance boost is particularly noteworthy for GPT4, because GPT4 by itself outperforms the additional opinions provided by IL models, while still benefiting from this method.

To explore the impact of single vs. multiple additional opinions, we also tested sampled top five additional opinions vs. one (see Table 1), and observed GPT4 with top 5 additional opinions reached the best Success Rate, Rewards and Precision across all groups (Table 1). From an intelligent agent perspective, awareness of an additional opinion, or even multiple distinct opinions, can be beneficial, as argued in \cite{additional_opinion}. This suggests that providing LLMs with one or a few additional opinions of reasonable quality can serve as a reference, resulting in a more informed decision. 

Out of curiosity, we also conducted one ablation study by providing random one additional opinion to GPT3.5. We observed the worst performance - the lowest reward (22.333) and an equivalently low success rate (0.060) as Rule based model. 

In Figure 2, we observe that Language Learning Models (LLMs) predominantly take in the additional opinion suggested by expert models, with GPT4 exhibiting the highest standard and the greatest proportion of disagreements. We consider any match among the top 5 additional opinions as being taken into account by the LLMs. Intriguingly, for GPT4, the ratio of considered opinions escalates from 0.549 to 0.602 as the number of opinions increases from 1 to 5. This trend could partially elucidate the disparity in the final outcomes of success rates.

\begin{figure}[hbt!] 
  \centering
  \includegraphics[width=\linewidth]{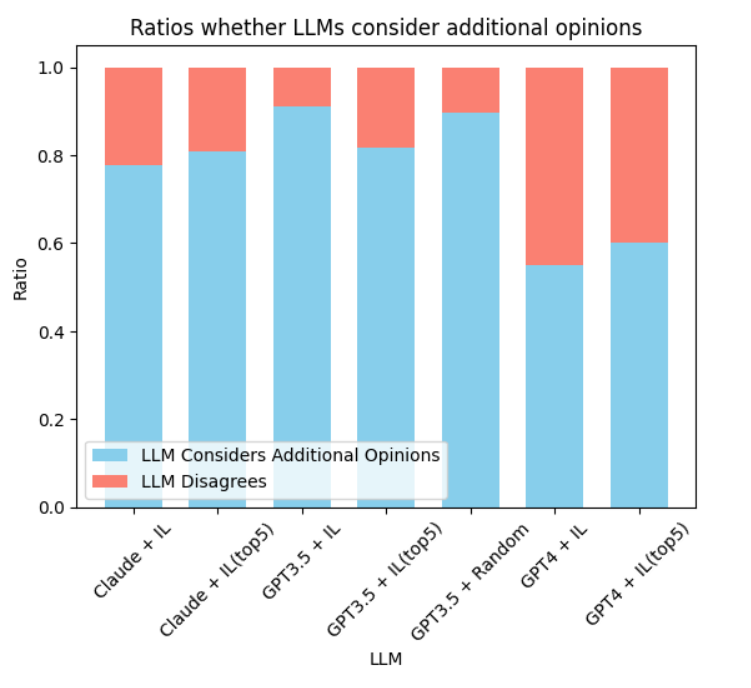}
  \caption{For Webshop, the ratios of LLMs considering or disagreeing with additional opinion provided by expert models. For the top 5 scenario, we consider it as an agreement if any additional opinion matches.}
  \label{fig:LLM_Consider_AO}
\end{figure}

\subsubsection{ALFWorld}
Taking inspiration from our previous Webshop experiment, we employed a similar paradigm to combine Large Language Models (LLMs) with Expert models in the context of the ALFWorld dataset. The fundamental premise remained analogous, but due to resource constraints and the extensive size of ALFWorld (134 unseen games), we confined our initial trial to a top 1 Imitation Learning (IL) opinion.

Our hypothesis received empirical confirmation from the data observed for GPT4 and Claude, as detailed in Table 2. The success rate showed slight improvements, although the original DAgger agent provided by ALFWorld was less effective for general tasks compared to specific ones, as demonstrated by the task-specific performance rates listed in Table 4 from the ALFWorld paper \cite{alfworld}, i.e in a disparate performance range, the DAgger agent excelled in the 'Cool and Place' task, achieving a remarkable 1.00 success rate. However, for the 'Pick Two and Place' task, its proficiency was considerably lower, with a success rate of only 0.24. Our investigation further revealed a substantial variance in efficiency concerning the number of steps taken by the ALFWorld IL model to complete tasks. This observation highlights an additional layer of complexity, suggesting that the IL's performance is not only variable in terms of task success rates, but also in the practical efficiency of task execution. This large variability in step numbers indicates that while the IL model may excel in some tasks, it can be markedly inefficient in others, further reinforcing the inherent limitations and variability of IL models in complex task simulations like ALFWorld. In terms of task completion, the IL's outcomes are governed by real-time rewards. As such, all completed tasks have to be executed correctly, as evidenced by 1.00 precision for the both IL with and without Beam Search. Conversely, the tasks left incomplete were typically ones where the agent had exhausted the available steps, with most of these marked by repeated and meaningless actions as presented in Appendix 2.

One of the standout observations from our study was GPT4's ability to integrate past memories, task descriptions, and environmental data to discern the pertinence of suggestions from the IL model. Even amidst noise, GPT4 demonstrated a robust capacity to differentiate beneficial from irrelevant advice, often confidently disregarding suggestions that were not beneficial, as illustrated in Appendix 3. Moreover, GPT4 was even able to extract values from the initial part of a repetitive action pattern suggested by the IL, underscoring its exceptional ability to distill useful information. Contrastingly, GPT3.5 was easily misled by irrelevant suggestions and frequently became entangled in the repetitive advice offered by the IL. Indeed, such confusion even compromised GPT3.5's capability to perform tasks that it could otherwise successfully accomplish independently, as detailed in Appendix 4. This highlights a stark divergence from the patterns observed with GPT4 under ALFWorld Context.

In a compelling revelation, this study demonstrated a marked difference between the contexts of the Webshop and ALFWorld experiments. The beneficial guidance provided by the Webshop's IL model effectively condensed the choice spaces for LLMs, contrasted with the repetitive and misleading advice offered by the IL in the ALFWorld context. Interesting discrepancies were also observed in how the LLMs disagreed with the IL's recommendations: Claude registered a disagreement rate of 0.814, GPT3.5 of 0.769, and GPT4, leading the pack, registered a rate of 0.854. This suggests an inherent capability within LLMs to filter out misleading suggestions. However, the extent to which this disagreement improved or impeded performance appeared to be context-dependent, highlighting the importance of discernment in processing the IL's advice. Claude and specially GPT4 showed remarkable adeptness at avoiding the pitfalls of misleading and repetitive advice. By contrast, GPT3.5 exhibited a clear shortfall in this respect, a performance echoed by its pairing with a random action in our Webshop ablation study. This underscores the importance of context when integrating IL models with LLMs and signals the need for careful evaluation when dealing with potentially misleading input, especially on an LLM like GPT3.5 which can get easily confused.

\begin{figure}[hbt!] 
  \centering
  \includegraphics[width=\linewidth]{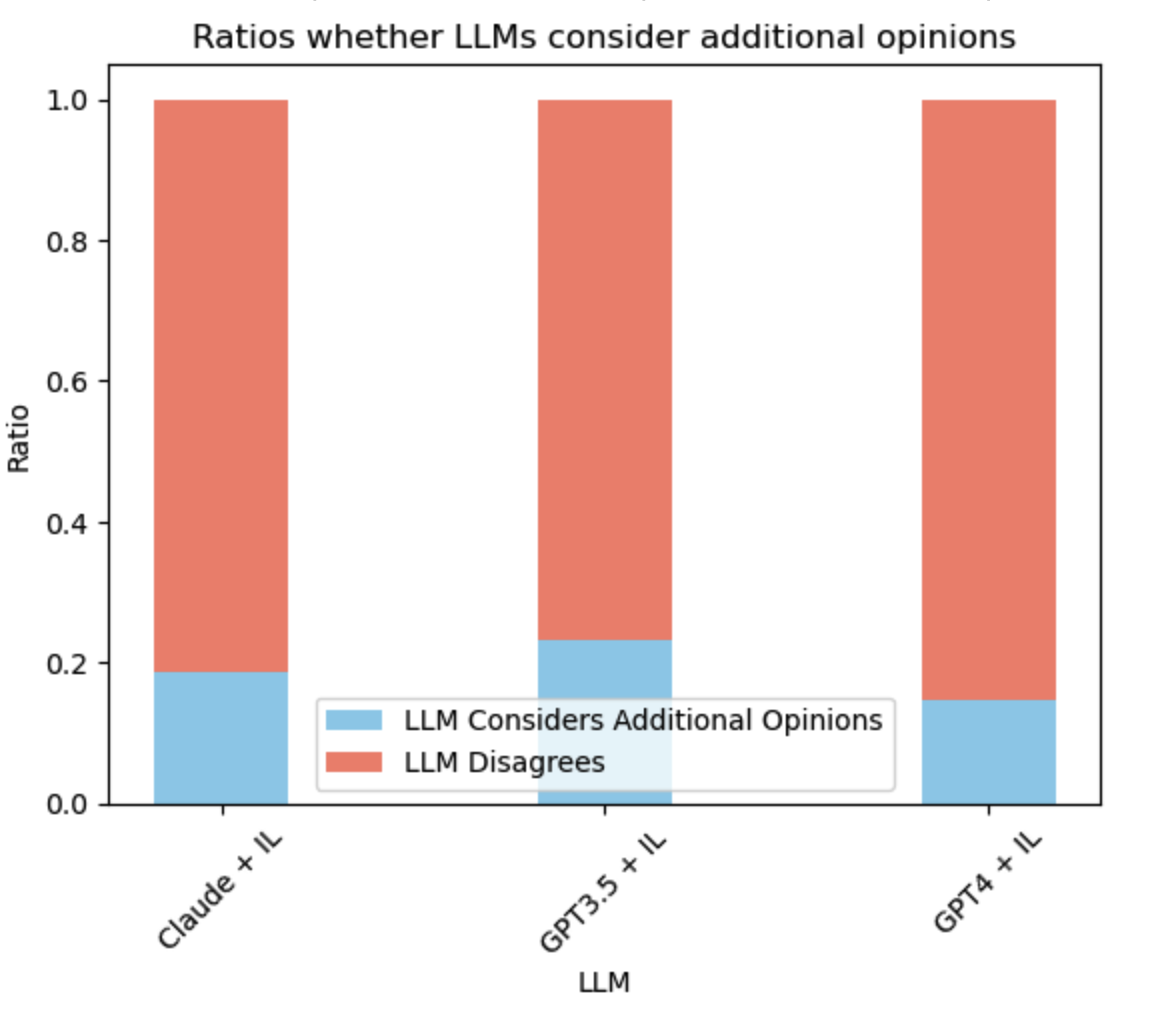}
  \caption{For ALFWorld,  the ratios of LLMs considering or disagreeing with additional opinion provided by expert models. For the top 1 case, we only redeem it as an agreement if the exact opinion matches.}
  \label{fig:LLM_Consider_ALFWorld}
\end{figure}

\subsection{Discussions}

Initially, Auto-GPT was conceptualized as an experimental idea rather than a robust workflow suitable for real-world applications. However, our research demonstrates otherwise. Auto-GPT not only proves its potential for practical use but also outperforms supervised state-of-the-art IL models with GPT4, signifying a shift in perspective towards this innovative approach.

In the current discourse, we posit that this additional opinion approach can readily find widespread adoption across diverse industries, given the existing prevalence of expert models such as recommendation systems and traditional natural language processing (NLP) services, inclusive of text classification models and the like. An immediate application for this methodology can be envisaged in leveraging LLMs for making definitive determinations and give explanations regarding the prioritization of items, such as movies or songs, to be displayed to the user. This is achievable by providing the selected top-k outputs derived from a supervised recommendation model to LLMs as Additional Opinions. 

It is crucial to note, however, that the two tasks we have chosen to benchmark in this research do not fully encapsulate the vast array of potential real-world scenarios. They serve merely as a starting point for the exploration of this idea. This is the inaugural instance where the concept of adapting Auto-GPT to handle complex tasks by introducing Additional Opinions has been proposed. This innovative approach opens new avenues for further research and development, potentially expanding the realm of practical applications for AI models and significantly impacting our understanding of complex decision-making mechanisms.

\section{Related Work}\label{sec:related_work}
Foundation models trained on self-supervision tasks have shown significant success in downstream tasks, particularly in few-shot and zero-shot settings \cite{bert, roberta, albert}. More recently, generative pre-trained foundation models have demonstrated impressive in-context learning abilities, allowing them to tackle decision-making tasks that require logic reasoning \cite{llm+p} and/or interactions with external APIs \cite{foundationsurvey, taskmatrix, gorilla}. However, adapting LLMs for decision-making tasks often involves non-trivial prompt design, memory retrieval mechanisms to dynamically construct the agent's context \cite{react, reflexion, generativeagent}, and sometimes model fine-tuning \cite{taskmatrix} to enhance its decision-making abilities.

Several techniques have been proposed to adapt Large Language Models (LLMs) for improved planning and reasoning. These include methods for enabling explicit Chain of Thought (CoT) thinking processes \cite{cot}, as well as prompting and decoding techniques aimed at enhancing the self-consistency of LLMs \cite{complexitybased, selfconsistency}. However, most of these techniques primarily focus on offline reasoning tasks that can be planned ahead, while their implications in online decision-making scenarios are rarely discussed.


\section{Conclusion}\label{sec:conclusion}

Our experimental results highlight the successful adaptation of the Auto-GPT styled agent to complex online decision-making tasks through straightforward prompt design, surpassing IL-based baseline models specifically designed for these tasks. Among the foundational LLMs powering Auto-GPT, GPT-4 demonstrates superior performance. Additionally, we introduce an innovative strategy of incorporating additional opinions from external expert models, further enhancing the decision-making capabilities of Auto-GPT styled agents, particularly benefiting GPT-4. Our Additional Opinions algorithm provides a lightweight supervised training approach for Auto-GPT styled agents, enabling improved performance without requiring extensive fine-tuning of the LLMs. We demonstrate the effectiveness and adaptability of this approach, especially for tasks with easily collectible training data for action policy. The code of this work is shared in: https://github.com/younghuman/LLMAgent


\bibliographystyle{ACM-Reference-Format}
\bibliography{bibliography}


\begin{thebibliography}{25}


\ifx \showCODEN    \undefined \def \showCODEN     #1{\unskip}     \fi
\ifx \showDOI      \undefined \def \showDOI       #1{#1}\fi
\ifx \showISBNx    \undefined \def \showISBNx     #1{\unskip}     \fi
\ifx \showISBNxiii \undefined \def \showISBNxiii  #1{\unskip}     \fi
\ifx \showISSN     \undefined \def \showISSN      #1{\unskip}     \fi
\ifx \showLCCN     \undefined \def \showLCCN      #1{\unskip}     \fi
\ifx \shownote     \undefined \def \shownote      #1{#1}          \fi
\ifx \showarticletitle \undefined \def \showarticletitle #1{#1}   \fi
\ifx \showURL      \undefined \def \showURL       {\relax}        \fi
\providecommand\bibfield[2]{#2}
\providecommand\bibinfo[2]{#2}
\providecommand\natexlab[1]{#1}
\providecommand\showeprint[2][]{arXiv:#2}

\bibitem[Bai et~al\mbox{.}(2022)]%
        {claude}
\bibfield{author}{\bibinfo{person}{Yuntao Bai}, \bibinfo{person}{Saurav
  Kadavath}, \bibinfo{person}{Sandipan Kundu}, \bibinfo{person}{Amanda Askell},
  \bibinfo{person}{Jackson Kernion}, \bibinfo{person}{Andy Jones},
  \bibinfo{person}{Anna Chen}, \bibinfo{person}{Anna Goldie},
  \bibinfo{person}{Azalia Mirhoseini}, {and} \bibinfo{person}{Cameron
  McKinnon}.} \bibinfo{year}{2022}\natexlab{}.
\newblock \showarticletitle{Constitutional AI: Harmlessness from AI Feedback}.
  In \bibinfo{booktitle}{\emph{arXiv:2212.08073}}. \bibinfo{publisher}{arXiv}.
\newblock
\urldef\tempurl%
\url{https://ar5iv.org/abs/2212.08073}
\showURL{%
\tempurl}


\bibitem[Bubeck et~al\mbox{.}(2023)]%
        {msgpt}
\bibfield{author}{\bibinfo{person}{Sébastien Bubeck}, \bibinfo{person}{Varun
  Chandrasekaran}, \bibinfo{person}{Ronen Eldan}, \bibinfo{person}{Johannes
  Gehrke}, \bibinfo{person}{Eric Horvitz}, \bibinfo{person}{Ece Kamar},
  \bibinfo{person}{Peter Lee}, \bibinfo{person}{Yin~Tat Lee},
  \bibinfo{person}{Yuanzhi Li}, \bibinfo{person}{Scott Lundberg},
  \bibinfo{person}{Harsha Nori}, \bibinfo{person}{Hamid Palangi},
  \bibinfo{person}{Marco~Tulio Ribeiro}, {and} \bibinfo{person}{Yi Zhang}.}
  \bibinfo{year}{2023}\natexlab{}.
\newblock \bibinfo{title}{Sparks of Artificial General Intelligence: Early
  experiments with GPT-4}.  (\bibinfo{date}{March} \bibinfo{year}{2023}).
\newblock
\urldef\tempurl%
\url{https://www.microsoft.com/en-us/research/publication/sparks-of-artificial-general-intelligence-early-experiments-with-gpt-4/}
\showURL{%
\tempurl}


\bibitem[Chowdhery et~al\mbox{.}(2022)]%
        {palm}
\bibfield{author}{\bibinfo{person}{Aakanksha Chowdhery},
  \bibinfo{person}{Sharan Narang}, \bibinfo{person}{Jacob Devlin},
  \bibinfo{person}{Maarten Bosma}, \bibinfo{person}{Gaurav Mishra},
  \bibinfo{person}{Adam Roberts}, \bibinfo{person}{Paul Barham},
  \bibinfo{person}{Hyung~Won Chung}, \bibinfo{person}{Charles Sutton},
  \bibinfo{person}{Sebastian Gehrmann}, {et~al\mbox{.}}}
  \bibinfo{year}{2022}\natexlab{}.
\newblock \showarticletitle{Palm: Scaling language modeling with pathways}.
\newblock \bibinfo{journal}{\emph{arXiv preprint arXiv:2204.02311}}
  (\bibinfo{year}{2022}).
\newblock


\bibitem[C{\^o}t{\'e} et~al\mbox{.}(2018)]%
        {textworld}
\bibfield{author}{\bibinfo{person}{Marc-Alexandre C{\^o}t{\'e}},
  \bibinfo{person}{{\'A}kos K{\'a}d{\'a}r}, \bibinfo{person}{Xingdi Yuan},
  \bibinfo{person}{Ben Kybartas}, \bibinfo{person}{Tavian Barnes},
  \bibinfo{person}{Emery Fine}, \bibinfo{person}{James Moore},
  \bibinfo{person}{Matthew Hausknecht}, \bibinfo{person}{Layla El~Asri},
  \bibinfo{person}{Mahmoud Adada}, {et~al\mbox{.}}}
  \bibinfo{year}{2018}\natexlab{}.
\newblock \showarticletitle{Textworld: A learning environment for text-based
  games}.
\newblock  (\bibinfo{year}{2018}), \bibinfo{pages}{41--75}.
\newblock


\bibitem[Devlin et~al\mbox{.}(2018)]%
        {bert}
\bibfield{author}{\bibinfo{person}{Jacob Devlin}, \bibinfo{person}{Ming-Wei
  Chang}, \bibinfo{person}{Kenton Lee}, {and} \bibinfo{person}{Kristina
  Toutanova}.} \bibinfo{year}{2018}\natexlab{}.
\newblock \showarticletitle{Bert: Pre-training of deep bidirectional
  transformers for language understanding}.
\newblock \bibinfo{journal}{\emph{arXiv preprint arXiv:1810.04805}}
  (\bibinfo{year}{2018}).
\newblock


\bibitem[Fu et~al\mbox{.}(2023)]%
        {complexitybased}
\bibfield{author}{\bibinfo{person}{Yao Fu}, \bibinfo{person}{Hao Peng},
  \bibinfo{person}{Ashish Sabharwal}, \bibinfo{person}{Peter Clark}, {and}
  \bibinfo{person}{Tushar Khot}.} \bibinfo{year}{2023}\natexlab{}.
\newblock \bibinfo{title}{Complexity-Based Prompting for Multi-Step Reasoning}.
\newblock
\newblock
\showeprint[arxiv]{2210.00720}~[cs.CL]


\bibitem[Lan et~al\mbox{.}(2019)]%
        {albert}
\bibfield{author}{\bibinfo{person}{Zhenzhong Lan}, \bibinfo{person}{Mingda
  Chen}, \bibinfo{person}{Sebastian Goodman}, \bibinfo{person}{Kevin Gimpel},
  \bibinfo{person}{Piyush Sharma}, {and} \bibinfo{person}{Radu Soricut}.}
  \bibinfo{year}{2019}\natexlab{}.
\newblock \showarticletitle{Albert: A lite bert for self-supervised learning of
  language representations}.
\newblock \bibinfo{journal}{\emph{arXiv preprint arXiv:1909.11942}}
  (\bibinfo{year}{2019}).
\newblock


\bibitem[Li et~al\mbox{.}(2018)]%
        {llm+p}
\bibfield{author}{\bibinfo{person}{Yang Li}, \bibinfo{person}{Zhi-Ping Cai},
  {and} \bibinfo{person}{Hong Xu}.} \bibinfo{year}{2018}\natexlab{}.
\newblock \showarticletitle{LLMP: exploiting LLDP for latency measurement in
  software-defined data center networks}.
\newblock \bibinfo{journal}{\emph{Journal of Computer Science and Technology}}
  \bibinfo{volume}{33} (\bibinfo{year}{2018}), \bibinfo{pages}{277--285}.
\newblock


\bibitem[Liang et~al\mbox{.}(2023)]%
        {taskmatrix}
\bibfield{author}{\bibinfo{person}{Yaobo Liang}, \bibinfo{person}{Chenfei Wu},
  \bibinfo{person}{Ting Song}, \bibinfo{person}{Wenshan Wu},
  \bibinfo{person}{Yan Xia}, \bibinfo{person}{Yu Liu}, \bibinfo{person}{Yang
  Ou}, \bibinfo{person}{Shuai Lu}, \bibinfo{person}{Lei Ji},
  \bibinfo{person}{Shaoguang Mao}, {et~al\mbox{.}}}
  \bibinfo{year}{2023}\natexlab{}.
\newblock \showarticletitle{Taskmatrix. ai: Completing tasks by connecting
  foundation models with millions of apis}.
\newblock \bibinfo{journal}{\emph{arXiv preprint arXiv:2303.16434}}
  (\bibinfo{year}{2023}).
\newblock


\bibitem[Liu et~al\mbox{.}(2019)]%
        {roberta}
\bibfield{author}{\bibinfo{person}{Yinhan Liu}, \bibinfo{person}{Myle Ott},
  \bibinfo{person}{Naman Goyal}, \bibinfo{person}{Jingfei Du},
  \bibinfo{person}{Mandar Joshi}, \bibinfo{person}{Danqi Chen},
  \bibinfo{person}{Omer Levy}, \bibinfo{person}{Mike Lewis},
  \bibinfo{person}{Luke Zettlemoyer}, {and} \bibinfo{person}{Veselin
  Stoyanov}.} \bibinfo{year}{2019}\natexlab{}.
\newblock \showarticletitle{Roberta: A robustly optimized bert pretraining
  approach}.
\newblock \bibinfo{journal}{\emph{arXiv preprint arXiv:1907.11692}}
  (\bibinfo{year}{2019}).
\newblock


\bibitem[OpenAI(2023)]%
        {gpt4}
\bibfield{author}{\bibinfo{person}{OpenAI}.} \bibinfo{year}{2023}\natexlab{}.
\newblock \bibinfo{title}{GPT-4 Technical Report}.
\newblock
\newblock
\showeprint[arxiv]{2303.08774}~[cs.CL]


\bibitem[Park et~al\mbox{.}(2023)]%
        {generativeagent}
\bibfield{author}{\bibinfo{person}{Joon~Sung Park}, \bibinfo{person}{Joseph~C.
  O'Brien}, \bibinfo{person}{Carrie Cai}, \bibinfo{person}{Meredith~Ringel
  Morris}, \bibinfo{person}{Percy Liang}, {and} \bibinfo{person}{Michael
  Bernstein}.} \bibinfo{year}{2023}\natexlab{}.
\newblock \showarticletitle{Generative Agents: Interactive Simulacra of Human
  Behavior}.
\newblock
\urldef\tempurl%
\url{https://arxiv.org/pdf/2304.03442.pdf}
\showURL{%
\tempurl}


\bibitem[Patil et~al\mbox{.}(2023)]%
        {gorilla}
\bibfield{author}{\bibinfo{person}{Shishir~G. Patil}, \bibinfo{person}{Tianjun
  Zhang}, \bibinfo{person}{Xin Wang}, {and} \bibinfo{person}{Joseph~E.
  Gonzalez}.} \bibinfo{year}{2023}\natexlab{}.
\newblock \bibinfo{title}{Gorilla: Large Language Model Connected with Massive
  APIs}.
\newblock
\newblock
\showeprint[arxiv]{2305.15334}~[cs.CL]


\bibitem[Peng et~al\mbox{.}(2022)]%
        {vicuna}
\bibfield{author}{\bibinfo{person}{Baolin Peng}, \bibinfo{person}{Chunyuan Li},
  \bibinfo{person}{Pengcheng He}, \bibinfo{person}{Michel Galley}, {and}
  \bibinfo{person}{Jianfeng Gao}.} \bibinfo{year}{2022}\natexlab{}.
\newblock \bibinfo{booktitle}{\emph{Instruction Tuning with GPT-4}}.
\newblock
\urldef\tempurl%
\url{https://ar5iv.org/abs/2304.03277}
\showURL{%
\tempurl}


\bibitem[Ross et~al\mbox{.}(2011)]%
        {ross2011reduction}
\bibfield{author}{\bibinfo{person}{Stephane Ross}, \bibinfo{person}{Geoffrey~J.
  Gordon}, {and} \bibinfo{person}{J.~Andrew Bagnell}.}
  \bibinfo{year}{2011}\natexlab{}.
\newblock \showarticletitle{A Reduction of Imitation Learning and Structured
  Prediction to No-Regret Online Learning}.
\newblock \bibinfo{journal}{\emph{arXiv preprint arXiv:1011.0686}}
  (\bibinfo{year}{2011}).
\newblock
\showeprint{1011.0686}


\bibitem[Shen et~al\mbox{.}(2023)]%
        {hugginggpt}
\bibfield{author}{\bibinfo{person}{Yongliang Shen}, \bibinfo{person}{Kaitao
  Song}, \bibinfo{person}{Xu Tan}, \bibinfo{person}{Dongsheng Li},
  \bibinfo{person}{Weiming Lu}, {and} \bibinfo{person}{Yueting Zhuang}.}
  \bibinfo{year}{2023}\natexlab{}.
\newblock \bibinfo{title}{HuggingGPT: Solving AI Tasks with ChatGPT and its
  Friends in HuggingFace}.
\newblock
\newblock
\showeprint[arxiv]{2303.17580}~[cs.CL]


\bibitem[Shinn et~al\mbox{.}(2023)]%
        {reflexion}
\bibfield{author}{\bibinfo{person}{Noah Shinn}, \bibinfo{person}{Beck Labash},
  {and} \bibinfo{person}{Ashwin Gopinath}.} \bibinfo{year}{2023}\natexlab{}.
\newblock \showarticletitle{Reflexion: an autonomous agent with dynamic memory
  and self-reflection}.
\newblock \bibinfo{journal}{\emph{arXiv preprint arXiv:2303.11366}}
  (\bibinfo{year}{2023}).
\newblock


\bibitem[Shridhar et~al\mbox{.}(2020)]%
        {alfred}
\bibfield{author}{\bibinfo{person}{Mohit Shridhar}, \bibinfo{person}{Jesse
  Thomason}, \bibinfo{person}{Daniel Gordon}, \bibinfo{person}{Yonatan Bisk},
  \bibinfo{person}{Winson Han}, \bibinfo{person}{Roozbeh Mottaghi},
  \bibinfo{person}{Luke Zettlemoyer}, {and} \bibinfo{person}{Dieter Fox}.}
  \bibinfo{year}{2020}\natexlab{}.
\newblock \showarticletitle{{ALFRED: A Benchmark for Interpreting Grounded
  Instructions for Everyday Tasks}}.
\newblock \bibinfo{journal}{\emph{arXiv preprint arXiv:1912.01734}}
  (\bibinfo{year}{2020}).
\newblock


\bibitem[Shridhar et~al\mbox{.}(2021)]%
        {alfworld}
\bibfield{author}{\bibinfo{person}{Mohit Shridhar}, \bibinfo{person}{Xingdi
  Yuan}, \bibinfo{person}{Marc-Alexandre C\^ot\'e}, \bibinfo{person}{Yonatan
  Bisk}, \bibinfo{person}{Adam Trischler}, {and} \bibinfo{person}{Matthew
  Hausknecht}.} \bibinfo{year}{2021}\natexlab{}.
\newblock \showarticletitle{AlfWorld: Aligning Text and Embodied Environments
  for Interactive Learning}.
\newblock \bibinfo{journal}{\emph{arXiv preprint arXiv:2010.03768}}
  (\bibinfo{year}{2021}).
\newblock


\bibitem[Wang et~al\mbox{.}(2023)]%
        {selfconsistency}
\bibfield{author}{\bibinfo{person}{Xuezhi Wang}, \bibinfo{person}{Jason Wei},
  \bibinfo{person}{Dale Schuurmans}, \bibinfo{person}{Quoc Le},
  \bibinfo{person}{Ed Chi}, \bibinfo{person}{Sharan Narang},
  \bibinfo{person}{Aakanksha Chowdhery}, {and} \bibinfo{person}{Denny Zhou}.}
  \bibinfo{year}{2023}\natexlab{}.
\newblock \bibinfo{title}{Self-Consistency Improves Chain of Thought Reasoning
  in Language Models}.
\newblock
\newblock
\showeprint[arxiv]{2203.11171}~[cs.CL]


\bibitem[Wei et~al\mbox{.}(2022)]%
        {cot}
\bibfield{author}{\bibinfo{person}{Jason Wei}, \bibinfo{person}{Xuezhi Wang},
  \bibinfo{person}{Dale Schuurmans}, \bibinfo{person}{Maarten Bosma},
  \bibinfo{person}{Ed Chi}, \bibinfo{person}{Quoc Le}, {and}
  \bibinfo{person}{Denny Zhou}.} \bibinfo{year}{2022}\natexlab{}.
\newblock \showarticletitle{Chain of thought prompting elicits reasoning in
  large language models}.
\newblock \bibinfo{journal}{\emph{arXiv preprint arXiv:2201.11903}}
  (\bibinfo{year}{2022}).
\newblock


\bibitem[Yang et~al\mbox{.}(2023)]%
        {foundationsurvey}
\bibfield{author}{\bibinfo{person}{Sherry Yang}, \bibinfo{person}{Ofir Nachum},
  \bibinfo{person}{Yilun Du}, \bibinfo{person}{Jason Wei},
  \bibinfo{person}{Pieter Abbeel}, {and} \bibinfo{person}{Dale Schuurmans}.}
  \bibinfo{year}{2023}\natexlab{}.
\newblock \showarticletitle{Foundation Models for Decision Making: Problems,
  Methods, and Opportunities}.
\newblock \bibinfo{journal}{\emph{arXiv e-prints}} (\bibinfo{year}{2023}),
  \bibinfo{pages}{arXiv--2303}.
\newblock


\bibitem[Yaniv(2004)]%
        {additional_opinion}
\bibfield{author}{\bibinfo{person}{Ilan Yaniv}.}
  \bibinfo{year}{2004}\natexlab{}.
\newblock \showarticletitle{The benefit of additional opinions}.
\newblock \bibinfo{journal}{\emph{Current directions in psychological science}}
  \bibinfo{volume}{13}, \bibinfo{number}{2} (\bibinfo{year}{2004}),
  \bibinfo{pages}{75--78}.
\newblock


\bibitem[Yao et~al\mbox{.}(2022a)]%
        {webshop}
\bibfield{author}{\bibinfo{person}{Shunyu Yao}, \bibinfo{person}{Howard Chen},
  \bibinfo{person}{John Yang}, {and} \bibinfo{person}{Karthik Narasimhan}.}
  \bibinfo{year}{2022}\natexlab{a}.
\newblock \showarticletitle{Webshop: Towards scalable real-world web
  interaction with grounded language agents}.
\newblock \bibinfo{journal}{\emph{arXiv preprint arXiv:2207.01206}}
  (\bibinfo{year}{2022}).
\newblock


\bibitem[Yao et~al\mbox{.}(2022b)]%
        {react}
\bibfield{author}{\bibinfo{person}{Shunyu Yao}, \bibinfo{person}{Jeffrey Zhao},
  \bibinfo{person}{Dian Yu}, \bibinfo{person}{Nan Du}, \bibinfo{person}{Izhak
  Shafran}, \bibinfo{person}{Karthik Narasimhan}, {and} \bibinfo{person}{Yuan
  Cao}.} \bibinfo{year}{2022}\natexlab{b}.
\newblock \showarticletitle{React: Synergizing reasoning and acting in language
  models}.
\newblock \bibinfo{journal}{\emph{arXiv preprint arXiv:2210.03629}}
  (\bibinfo{year}{2022}).
\newblock


\end{thebibliography}

\appendix
\section{Appendix}\label{sec:appendix}
\begin{figure*}
\subsection{APPENDIX 1. Webshop: Evidence of GPT-4 considering additional opinion.}
\subsubsection{Example 1. Webshop: AutoGPT(GPT-4) disagreed with the Additional Opinion}
\begin{verbatim}[breaklines]
Command search returned: =Observation=
Instruction:
i want a pink niuta 2 pack hair towel wrap for dry hair, and price lower than 20.00 dollars
[button] Back to Search [button_]
Page 1 (Total results: 50)
[button] Next > [button_]

[button] B08G14B779 [button_]
NIUTA 2 Pack Hair Towel Wrap, Microfiber Quick Drying Hair Towels, Super Absorbent Quick Dry Hair Towel, 
Wrapped Bath Cap (Pink+Light Blue)
$4.99

[button] B092QTHX2Z [button_]
Microfiber Hair Towel Wrap POPCHOSE Ultra Absorbent, Fast Drying Hair Turban Soft, No Frizz Hair Wrap Towels for Women Wet Hair, 
Curly, Longer, Thicker Hair…
$8.99

[button] B08NX5S6S2 [button_]
Microfiber Hair Towel Wrap for Women Super Absorbent Quick Dry Hair Towels with Button Design for Drying Curly&Thick Hair 2 
Pack(Purple+Camel)
$8.99

[button] B09KKCQ7FK [button_]
VOLO Hero Microfiber Hair Towel, Super Absorbent, Ultra-Soft, Fast Drying | Reduce Dry Time by 50%, Large, Premium Wrap Towel, 
Sustainable Packaging, Anti-Frizz | Luna Gray (2 Pack)
$70.0

[button] B083FPQ77W [button_]
Microfiber Hair Towels Super Absorbent Hair Drying Towel Turban for Women and Girls Quick Magic Hair Dry Hat Hair Towel Wrap Bathing
Wrapped Cap 2 Packs (Pink+Grey)
$13.99

[button] B09KK4GZW1 [button_]
VOLO Hero Microfiber Hair Towel | Super Absorbent, Ultra-Soft, Fast Drying | Reduce Dry Time by 50% | Large, Premium Wrap Towel for 
All Hair Types | Anti-Frizz, Anti-Breakage | Luna Gray (2 Pack)
$78.0

[button] B074T227KV [button_]
Microfiber Hair Towel, 2 Pack Dry Hair Towel Twist Wrap Absorbent Quickly Dry Hair Towel for Kids and Women (Pink+Purple)
$10.97

[button] B08CZBMHDR [button_]
Hair Towel Wrap, [2 Pack] Microfiber Quick Dry Hair Turban Wrap - Super Absorbent,Microfiber Hair Towel with Button,Quick Dry 
Hair Towel for Women,Hair Turbans for Wet Hair,Wrapped Bath Cap(Pink Blue)
$10.99

[button] B09FB25N5V [button_]
SYMGILA Microfiber Hair Towel Wrap, Hair Turbans for Wet Hair, Drying Hair Wrap Towels for Curly Hair Women Anti Frizz 2 Pack 
(Blue/Beige)
$100.0

[button] B08YMYXBK8 [button_]
VOLO Hero Microfiber Hair Towel, Super Absorbent, Ultra-Soft, Fast Drying | Reduce Dry Time by 50%, Large, Premium Wrap Towel,
Sustainable Packaging, Anti-Frizz, Anti-Breakage, Hands-Free (Luna Gray)
$34.99



loop number:2
Here's one suggestion for the command: click with 'back to search'.
Please use this suggestion as a reference and make your own judgement. Determine which next command to use, and respond 
using the JSON format specified above:
{
    "thoughts": {
        "text": "I found the matching item with a price lower than $20.",
        "reasoning": "The first item (B08G14B779) matches the description and has a price of $4.99, which is lower than $20.",
        "plan": "- click on the matching item (B08G14B779)\n- review the item details\n- make the purchase",
        "criticism": "Clicking 'back to search' is not necessary since I already found the matching item."
    },
    "command": {
        "name": "click",
        "args": {
            "tool_input": "B08G14B779"
        }
    }
}

\end{verbatim}
\end{figure*}

\begin{figure*}
\subsubsection{Example 2. Webshop: AutoGPT(GPT-4) agreed with the Additional Opinion}
\begin{verbatim}[breaklines,fontsize=\footnotesize]{text}
Command search returned: =Observation=
Instruction:
i want to buy a folding storage box ottoman which i can easily install and has faux leather, the size of it should be 60x40x40cm,
and price lower than 160.00 dollars
[button] Back to Search [button_]
Page 1 (Total results: 50)
[button] Next > [button_]

[button] B09LYXGGDD [button_]
Storage Stool Faux Leather Pouffe Folding Storage Ottoman Footstool Box Toy Chest with Lid, Foldable Foot Stool Seat for Living Room,
Bedroom (Pink,15.7x15.7x15.7in) Footstool Sofa Stool
$274.66

[button] B09LYZ12DH [button_]
Storage stool Folding Ottoman Storage Bench Cube Hold up 881lbs Faux Leather Long Chest with Memory Foam Seat Footrest Padded 
Upholstered Stool for Bedroom Box Bed Coffee Table Rectangular Purple Foot
$344.02

[button] B09NXSRVSK [button_]
Storage stool Folding Ottoman Storage Bench Cube Hold up 881lbs Faux Leather Long Chest with Memory Foam Seat Footrest Padded 
Upholstered Stool for Bedroom Box Bed Coffee Table Rectangular Brown (Size
$408.25

[button] B09NXP95GC [button_]
Storage stool Storage Ottoman Bench, Foldable Footrest Shoe Bench with Large Space, End of Bed Storage Seat, Storage Stool Box Toy 
Chest Support 881lbs, Faux Leather fabric (Colourl:Orange,Size:40x40x
$139.42

[button] B09NTCCVGX [button_]
Ottoman Footstools Folding Storage Box with Lid, Large Storage Seat Faux Leather Footstool Toy Storage Box, Yellow, 40x40x40cm (Color :
Yellow, Size : 40X40X40cm)
$149.97

[button] B09NXRK5MX [button_]
Storage stool Faux Leather Folding Organizer Storage Ottoman Bench Cube Foot Stool, Footrest Step Stool for Living Room, Bedroom,
Office, Garden, Traveling, Fishing, Camping 15.7"x15.7"x15.7" (Purple)
$150.01

[button] B09NTFLKXR [button_]
Folding Storage Ottoman Bench,Storage Chest Footrest Padded Seat,Faux Leather Ottoman Foot Rest Stool for Bedroom(Colour: Black,Size:
40x40x40cm/15.7x15.7x15.7in) (Color : Black, Size : 60X40X40cm)
$181.34

[button] B09MYV58YY [button_]
QTQHOME Rectangular Faux Leather Storage Ottoman Bench,Modern Tufted Storage Bench with Metal Leg,Upholstered Shoe Bench for Living 
Room Entryway Bedroom-Orange 60x40x40cm(24x16x16inch)
$100.0

[button] B08F3LZKSB [button_]
YZZSJC Storage Stool Foot Stool Rest Ottoman Storage Boxes, Cloth Shoe Changing Stool Storage Box Minimalist Clothing Store Sofa Stool
(Color : Light Blue, Size : 60x40x40cm)
$264.04

[button] B089D5K8LY [button_]
XMZDDZ Ottoman Storage Seat Bench,Faux Leather Tufted Storage Bench Footrest Stool with Padded Seat,Rectangular Bed Bench for Entryway
Bedroom Pink 60x40x40cm(24x16x16in)
$100.0

loop number:2
Here's one suggestion for the command: click with 'b09nxp95gc'.
Please use this suggestion as a reference and make your own judgement. Determine which next command to use, and respond using the JSON 
format specified above:
{
    "thoughts": {
        "text": "I need to click on the item B09NXP95GC.",
        "reasoning": "The item B09NXP95GC matches the size, material, and price requirements.",
        "plan": "- click on the item\n- review the item details\n- make the purchase",
        "criticism": "N/A"
    },
    "command": {
        "name": "click",
        "args": {
            "tool_input": "B09NXP95GC"
        }
    }
}
\end{verbatim}
\end{figure*}

\begin{figure*}
\subsection{APPENDIX 2. ALFWorld: Evidence of ALFWorld DAgger Agent }
\subsubsection{Example 1. ALFWorld: DAgger Agent repeated actions for "pick and place" task}
\begin{verbatim}[breaklines,fontsize=\footnotesize]{text}

pick_and_place_simple-PepperShaker-None-Drawer-10
look | go to cabinet 4 | go to cabinet 4 | go to cabinet 4 | go to cabinet 1 | go to cabinet 4 | go to cabinet 4 | go to cabinet 4 | go to 
cabinet 4 | go to cabinet 4 | go to cabinet 3 | take saltshaker 1 from cabinet 3 | go to drawer 3 | open drawer 3 | put saltshaker 1 in/on drawer
3 | close drawer 3 | go to cabinet 2 | open cabinet 2 | close cabinet 2 | go to cabinet 2 | go to cabinet 2 | go to cabinet 2 | go to cabinet 2 | 
go to cabinet 2 | go to cabinet 2 | go to cabinet 2 | go to cabinet 2 | go to cabinet 2 | go to cabinet 2 | go to cabinet 2 | go to cabinet 2 | 
go to cabinet 2 | go to cabinet 2 | go to cabinet 2 | go to cabinet 2 | go to cabinet 2 | go to cabinet 2 | go to cabinet 2 | go to cabinet 2 |
go to cabinet 2 | go to cabinet 2 | go to cabinet 2 | go to cabinet 2 | go to cabinet 2 | go to cabinet 2 | go to cabinet 2 | go to cabinet 2 | 
go to cabinet 2 | go to cabinet 2 | go to cabinet 2

pick_and_place_simple-Watch-None-Safe-219
look | go to shelf 2 | go to shelf 1 | go to shelf 1 | go to shelf 1 | go to shelf 12 | go to shelf 12 | go to shelf 12 | go to shelf 12 | go to 
shelf 12 | go to shelf 12 | go to shelf 12 | go to shelf 12 | go to shelf 12 | go to shelf 12 | go to shelf 12 | go to shelf 12 | go to shelf 12 
| go to shelf 12 | go to shelf 12 | go to shelf 12 | go to shelf 12 | go to shelf 12 | go to shelf 12 | go to shelf 12 | go to shelf 12 | go to 
shelf 12 | go to shelf 12 | go to shelf 12 | go to shelf 12 | go to shelf 12 | go to shelf 12 | go to shelf 12 | go to shelf 12 | go to shelf 12 
| go to shelf 12 | go to shelf 12 | go to shelf 12 | go to shelf 12 | go to shelf 12 | go to shelf 12 | go to shelf 12 | go to shelf 12 | go to 
shelf 12 | go to shelf 12 | go to shelf 12 | go to shelf 12 | go to shelf 12 | go to shelf 12 | go to shelf 12

pick_and_place_simple-Mug-None-Desk-308
look | go to shelf 5 | go to shelf 2 | go to shelf 2 | go to shelf 5 | go to shelf 5 | go to shelf 5 | go to shelf 5 | go to shelf 5 | go to 
shelf 5 | go to shelf 5 | go to shelf 5 | go to shelf 5 | go to shelf 5 | go to shelf 5 | go to shelf 5 | go to shelf 5 | go to shelf 5 | go to 
shelf 5 | go to shelf 5 | go to shelf 5 | go to shelf 5 | go to shelf 5 | go to shelf 5 | go to shelf 5 | go to shelf 5 | go to shelf 5 | go to 
shelf 5 | go to shelf 5 | go to shelf 5 | go to shelf 5 | go to shelf 5 | go to shelf 5 | go to shelf 5 | go to shelf 5 | go to shelf 5 | go to 
shelf 5 | go to shelf 5 | go to shelf 5 | go to shelf 5 | go to shelf 5 | go to shelf 5 | go to shelf 5 | go to shelf 5 | go to shelf 5 | go to 
shelf 5 | go to shelf 5 | go to shelf 5 | go to shelf 5 | go to shelf 5

\end{verbatim}
\end{figure*}

\begin{figure*}
\subsubsection{Example 2. ALFWorld: DAgger Agent effective action for "pick clean then place" task}
\begin{verbatim}[breaklines,fontsize=\footnotesize]{text}

pick_clean_then_place_in_recep-Cloth-None-CounterTop-424
look | go to cabinet 1 | open cabinet 1 | close cabinet 1 | go to countertop 1 | take cloth 3 from countertop 1 | go to sinkbasin 1 | clean cloth
3 with sinkbasin 1 | go to countertop 1 | put cloth 3 in/on countertop 1 | --

pick_clean_then_place_in_recep-Pan-None-CounterTop-1
look | go to countertop 1 | take pan 1 from countertop 1 | go to sinkbasin 1 | clean pan 1 with sinkbasin 1 | go to countertop 3 | put pan 1 
in/on countertop 3 | --

pick_clean_then_place_in_recep-Spatula-None-Drawer-10
look | go to countertop 3 | take spatula 1 from countertop 3 | go to sinkbasin 1 | clean spatula 1 with sinkbasin 1 | go to drawer 2 | open 
drawer 2 | put spatula 1 in/on drawer 2 | --

\end{verbatim}
\end{figure*}

\begin{figure*}
\subsection{APPENDIX 3. ALFWorld: Evidence of GPT-4 considering additional opinion.}
\subsubsection{Example 1. ALFWorld: AutoGPT(GPT-4) disagreed with the Additional Opinion}
\begin{verbatim}[breaklines,fontsize=\footnotesize]{text}

Instruction:
You are in the middle of a room. Looking quickly around you, you see a cabinet 4, a cabinet 3, a cabinet 2, a cabinet 1, a countertop 1, a 
garbagecan 1, a handtowelholder 2, a handtowelholder 1, a sinkbasin 2, a sinkbasin 1, a toilet 1, a toiletpaperhanger 1, and a towelholder 1.Your
task is to: clean some soapbar and put it in countertop.
After you go to countertop 1, current observation is: On the countertop 1, you see a soapbar 1, and a spraybottle 1.
Available Actions are:
action] examine countertop 1 [action_]
[action] go to cabinet 1 [action_]
[action] go to cabinet 2 [action_]
[action] go to cabinet 3 [action_]
[action] go to cabinet 4 [action_]
[action] go to garbagecan 1 [action_]
[action] go to handtowelholder 1 [action_]
[action] go to handtowelholder 2 [action_]
[action] go to sinkbasin 1 [action_]
[action] go to sinkbasin 2 [action_]
[action] go to toilet 1 [action_]
[action] go to toiletpaperhanger 1 [action_]
[action] go to towelholder 1 [action_]
[action] inventory [action_]
[action] look [action_]
[action] take soapbar 1 from countertop 1 [action_]
[action] take spraybottle 1 from countertop 1 [action_]

HumanMessage(content='
Here's one suggestion for the command: ['put handtowel 1 in/on countertop 1'].
Please use this suggestion as a reference and make your own judgement. Determine which next command to use, and respond using the JSON format 
specified above:')

AIMessage(content='
{
    "thoughts": {
        "text": "...",
        "reasoning": "The suggestion to put handtowel 1 in/on countertop 1 is not relevant to the task. I found the soapbar on countertop 1,
        so I should take it and proceed with cleaning it.",
        "plan": "- Take the soapbar\\n- Clean the soapbar\\n- Put the soapbar on the countertop",
        "criticism": "N/A."
    },
    "command": {
        "name": "alfworld_action",
        "args": {
            "tool_input": "take soapbar 1 from countertop 1"
        }
    }
}')


\end{verbatim}
\end{figure*}

\begin{figure*}
\subsubsection{Example 2. ALFWorld: AutoGPT(GPT-4) agreed with the Additional Opinion}
\begin{verbatim}[breaklines,fontsize=\footnotesize]{text}
Instruction:
You are in the middle of a room. Looking quickly around you, you see a cabinet 4, a cabinet 3, a cabinet 2, a cabinet 1, a countertop 1, a 
garbagecan 1, a handtowelholder 2, a handtowelholder 1, a sinkbasin 2, a sinkbasin 1, a toilet 1, toiletpaperhanger 1, and a towelholder 1. Your 
task is to: clean some soapbar and put it in countertop.
After you open cabinet 4, current observation is: Nothing happens.
Available Actions are:
[action] close cabinet 3 [action_]
[action] examine cabinet 3 [action_]
[action] go to cabinet 1 [action_]
[action] go to cabinet 2 [action_]
[action] go to cabinet 4 [action_]
[action] go to countertop 1 [action_]
[action] go to garbagecan 1 [action_]
[action] go to handtowelholder 1 [action_]
[action] go to handtowelholder 2 [action_]
[action] go to sinkbasin 1 [action_]
[action] go to sinkbasin 2 [action_]
[action] go to toilet 1 [action_]
[action] go to toiletpaperhanger 1 [action_]
[action] go to towelholder 1 [action_]
[action] inventory [action_]
[action] look [action_]
[action] take candle 1 from cabinet 3 [action_]
[action] take toiletpaper 3 from cabinet 3 [action_] 
 
HumanMessage(content="
Here's one suggestion for the command: ['go to countertop 1'].
Please use this suggestion as a reference and make your own judgement. Determine which next command to use, and respond using the JSON format
specified above:")

AIMessage(content='{
    "thoughts": {
        "text": "...",
        "reasoning": "I have checked all the cabinets and didn\'t find a soapbar. I will now go to countertop 1 to check if there\'s a soapbar
        there.",
        "plan": "- Find and take the soapbar\\n- Clean the soapbar\\n- Put the soapbar on the countertop",
        "criticism": "N/A."
    },
    "command": {
        "name": "alfworld_action",
        "args": {
            "tool_input": "go to countertop 1"
        }
    }
}')




\end{verbatim}
\end{figure*}

\begin{figure*}
\subsection{APPENDIX 4. ALFWorld: Evidence of GPT-3.5 confused by additional opinion.}
\subsubsection{Example 1. ALFWorld: AutoGPT(GPT-3.5) performed the task by it's own}
\begin{verbatim}[breaklines,fontsize=\footnotesize]{text}

Instruction:
You are in the middle of a room. Looking quickly around you, you see a bed 1, a desk 2, a desk 1, a drawer 6, a drawer 5, a drawer 4, a drawer 3,
a drawer 2, a drawer 1, a garbagecan 1, a laundryhamper 1, a safe 1, a shelf 6, a shelf 5, a shelf 4, a shelf 3, a shelf 2, and a shelf 1.Your 
task is to: examine the mug with the desklamp.\nAfter you go to shelf 2, current observation is: On the shelf 2, you see a creditcard 2, and a 
pen 2.
Available Actions are:
[action] examine shelf 2 [action_]
[action] go to bed 1 [action_]
[action] go to desk 1 [action_]
[action] go to desk 2 [action_]
[action] go to drawer 1 [action_]
[action] go to drawer 2 [action_]
[action] go to drawer 3 [action_]
[action] go to drawer 4 [action_]
[action] go to drawer 5 [action_]
[action] go to drawer 6 [action_]
[action] go to garbagecan 1 [action_]
[action] go to laundryhamper 1 [action_]
[action] go to safe 1 [action_]
[action] go to shelf 1 [action_]
[action] go to shelf 3 [action_]
[action] go to shelf 4 [action_]
[action] go to shelf 5 [action_]
[action] go to shelf 6 [action_]
[action] inventory [action_]
[action] look [action_]
[action] take creditcard 2 from shelf 2 [action_]
[action] take pen 2 from shelf 2 [action_]

HumanMessage(content='Determine which next command to use, and respond using the JSON format specified above:')
AIMessage(content='
{
    "thoughts": {
        "text": "...",
        "reasoning": "I need to find a mug and a desklamp. A mug is more likely to appear in shelf 1-6, desk 1-2, drawer 1-6. A desklamp is more
        likely to appear in desk 1-2, shelf 1-6. I can check one by one, starting with shelf 3.",
        "plan": "- Find a mug and a desklamp\\n- Examine the mug with the desklamp",
        "criticism": "N/A."
    },
    "command": {
        "name": "alfworld_action",
        "args": {
            "tool_input": "go to shelf 3"
        }
    }
}')
\end{verbatim}
\end{figure*}

\begin{figure*}
\subsubsection{Example 2. ALFWorld: AutoGPT(GPT-3.5) confused by IL repeated opinion}
\begin{verbatim}[breaklines,fontsize=\footnotesize]{text}
Instruction:
You are in the middle of a room. Looking quickly around you, you see a bed 1, a desk 2, a desk 1, a drawer 6, a drawer 5, a drawer 4, a drawer 3,
a drawer 2, a drawer 1, a garbagecan 1, a laundryhamper 1, a safe 1, a shelf 6, a shelf 5, a shelf 4, a shelf 3, a shelf 2, and a shelf 1.Your 
task is to: examine the mug with the desklamp.
After you go to shelf 2, current observation is: Nothing happens.
Available Actions are:
[action] examine shelf 2 [action_]
[action] go to bed 1 [action_]
[action] go to desk 1 [action_]
[action] go to desk 2 [action_]
[action] go to drawer 1 [action_]
[action] go to drawer 2 [action_]
[action] go to drawer 3 [action_]
[action] go to drawer 4 [action_]
[action] go to drawer 5 [action_]
[action] go to drawer 6 [action_]
[action] go to garbagecan 1 [action_]
[action] go to laundryhamper 1 [action_]
[action] go to safe 1 [action_]
[action] go to shelf 1 [action_]
[action] go to shelf 3 [action_]
[action] go to shelf 4 [action_]
[action] go to shelf 5 [action_]
[action] go to shelf 6 [action_]
[action] inventory [action_]
[action] look [action_]
[action] take creditcard 2 from shelf 2 [action_]
[action] take pen 2 from shelf 2 [action_]

HumanMessage(content="Here's one suggestion for the command: ['go to shelf 2'].
Please use this suggestion as a reference and make your own judgement. Determine which next command to use, and respond using the JSON format 
specified above:")

AIMessage(content='{
    "thoughts": {
        "text": "I will use the \'alfworld_action\' command to examine shelf 2 again.",
        "reasoning": "Since I did not find the desklamp on shelf 2, I need to examine it again to make sure I didn\'t miss               anything.",
        "plan": "- Use the \'alfworld_action\' command to go to shelf 2.\\\\n- Examine the items on the shelf again to  find the desklamp.
        \\\\n- Once I find the desklamp, I can use it to examine the mug.",
        "criticism": "I should have been more thorough in my examination of shelf 2 the first time."
        },
    "command": {
        "name": "alfworld_action",
        "args": {
            "tool_input": "go to shelf 2"
        }
    }
})
\end{verbatim}
\end{figure*}

\end{document}